\documentclass[11pt]{article}

\usepackage[final]{acl}

\usepackage{times}
\usepackage{latexsym}
\usepackage[table]{xcolor}
\usepackage[T1]{fontenc}
\usepackage{stfloats}  
\usepackage[utf8]{inputenc}

\usepackage{microtype}

\usepackage{inconsolata}
\usepackage{xcolor}
\definecolor{MidnightBlue}{HTML}{191970}
\usepackage{graphicx}
\usepackage[utf8]{inputenc} 
\usepackage[T1]{fontenc}    
\usepackage{hyperref}       
\usepackage{url}            
\usepackage{booktabs}       
\usepackage{amsfonts}       
\usepackage{nicefrac}       
\usepackage{microtype}      
\usepackage{xcolor}         
\usepackage{enumitem}
\usepackage{graphicx}
\usepackage[table]{xcolor}
\usepackage{booktabs}
\usepackage{makecell}
\usepackage{multirow}
\PassOptionsToPackage{numbers, compress}{natbib}
\definecolor{rowgray}{gray}{0.92} 
\usepackage{arydshln}
\usepackage{amsmath}  
\usepackage{amssymb}  
\usepackage{arydshln}

\usepackage[most]{tcolorbox} 

\tcbset{
  aibox/.style={
    width=\linewidth,
    top=8pt,
    bottom=4pt,
    colback=blue!6!white,   
    colframe=black,         
    colbacktitle=black,     
    coltitle=white,         
    fonttitle=\bfseries,    
    enhanced,
    center,
    attach boxed title to top left={yshift=-0.1in,xshift=0.15in},
    boxed title style={boxrule=0pt,colframe=white,},
    sharp corners,          
    breakable,              
  }
}

\NewDocumentCommand{\hongru}
{ mO{} }{\textcolor{red}{\textsuperscript{\textit{Hongru}}\textsf{\textbf{\small[#1]}}}}

\newtcolorbox{AIbox}[2][]{aibox,title=#2,#1}
\title{Demystifying On-Policy Distillation: Roles, Pathologies, and Regulations}

\author{
Rui Wang\textsuperscript{\dag},  Hongru Wang\textsuperscript{\dag}, Yi Chen, Boyang Xue\textsuperscript{\dag}\\
\textbf{Tianqing Fang\textsuperscript{\ddag*}, \textbf{Wenhao Yu\textsuperscript{\ddag}}, Kam-Fai Wong\textsuperscript{\dag*}} \\
\textsuperscript{\dag}The Chinese University of Hong Kong 
\textsuperscript{\ddag}Tencent AI Lab\\
\url{ruiwangnlp@outlook.com}\quad \url{fangtq229@gmail.com}\quad \url{kfwong@cuhk.edu.hk}\\
}

\begin{document}
\maketitle
\begin{abstract}

On-policy distillation (OPD) has become a key paradigm in LLM post-training, yet its training dynamics remain poorly understood. 
We present a systematic study examining the \textbf{role}, \textbf{pathologies}, and \textbf{regulations} of OPD. 
We first clarify the {\textit{role}} of OPD as an \textit{exploration catalyst}: it steers the student toward correct reasoning paths via dense token-level guidance, without expanding capability ceiling.
We confirm this by showing that prompt diversity matters more than per-problem sampling numbers, and critically, that the effectiveness of OPD hinges entirely on the quality of its guiding signal. 
This dependency exposes two {\textit{pathologies}} that derail exploration. 
The { Student-Teacher Mismatch} occurs when a large teacher-student distributional gap causes the guiding signal to misalign with task correctness, steering exploration in counterproductive directions. 
{Length Exploitation} arises when the aggregated token-level objective creates length-dependent shortcuts, allowing the student to game the reward landscape through response truncation or redundant padding, exploring degenerate length modes rather than reasoning strategies. 
To tame these pathologies, we investigate lightweight signal {\textit{regulations}}: \textit{advantage clipping} and \textit{log-scale compression}, ensuring exploration is guided by faithful signals. 
Experiments across seven benchmarks demonstrate that these regulations eliminate length exploitation and enable effective distillation, stably surpassing naive OPD and RLVR baselines, thereby confirming that well-regulated signal quality, rather than mere teacher scale, governs successful exploration in OPD. 

\end{abstract}

\section{Introduction}
\begin{figure}
    \centering
    \includegraphics[width=1\linewidth]{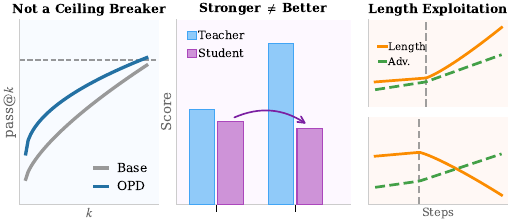}
    \vspace{-25pt}
    \caption{\textbf{Left}: OPD accelerates exploration but cannot exceed the model ceiling. \textbf{Middle}: a stronger teacher does not guarantee a better student. \textbf{Right}: advantage rises while length explodes or crashes.}
    \label{fig:teaser-toy}
    \vspace{-15pt}
\end{figure}


\definecolor{rolebg}{HTML}{EFF6FF}
\definecolor{roleframe}{HTML}{4A90D9}
\definecolor{pathbg}{HTML}{FEF3F2}
\definecolor{pathframe}{HTML}{D94F4F}
\definecolor{regbg}{HTML}{EEFBF3}
\definecolor{regframe}{HTML}{27AE60}

\begin{figure*}[t]
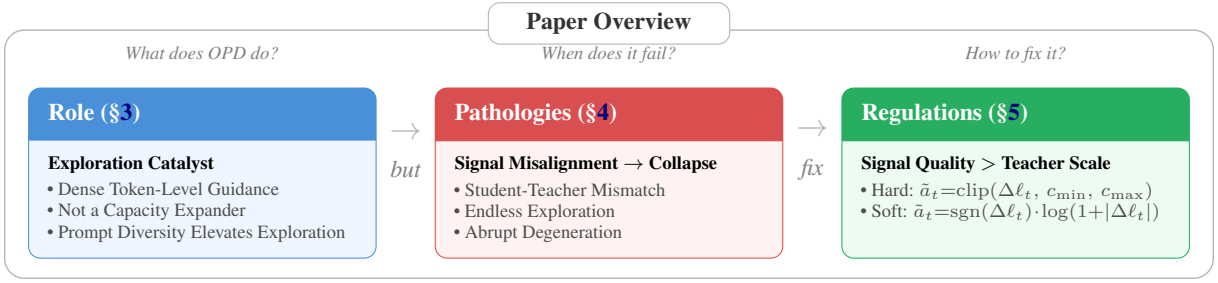

\begin{tcolorbox}[
  enhanced,
  colback=white,
  colframe=black!30,
  boxrule=0.5pt,
  arc=2.5mm,
  left=2mm, right=2mm,
  top=0.5mm, bottom=1.5mm,
  fonttitle=\small\bfseries,
  title={Paper Overview},
  coltitle=black!80,
  colbacktitle=white,
  attach boxed title to top center={yshift=-2mm},
  boxed title style={colframe=black!30, boxrule=0.4pt, arc=1.5mm, left=4mm, right=4mm},
]
\vspace{0.5mm}
\noindent
\begin{minipage}[c]{0.30\linewidth}
\centering{\scriptsize\color{black!50}\textit{What does OPD do?}}\\[-2.5pt]
\begin{tcolorbox}[
  enhanced,
  colback=rolebg,
  colframe=roleframe,
  fonttitle=\bfseries\footnotesize,
  title={Role (\S\ref{sec:role})},
  coltitle=white,
  colbacktitle=roleframe,
  toptitle=1mm, bottomtitle=1mm,
  boxrule=0.4pt,
  arc=1.5mm,
  top=1mm, bottom=1mm, left=1.5mm, right=1.5mm,
  fontupper=\scriptsize,
  equal height group=overviewpillar,
]
\textbf{Exploration Catalyst}\\[2pt]
{\scriptsize\color{black!70}%
\textbullet~Dense Token-Level Guidance\\
\textbullet~Not a Capacity Expander\\
\textbullet~Prompt Diversity Elevates Exploration}
\end{tcolorbox}
\end{minipage}%
\hfill
\begin{minipage}[c]{0.03\linewidth}
\centering
{\color{black!30}\large$\rightarrow$}\\[1pt]
{\color{black!50}\footnotesize\itshape but}
\end{minipage}%
\hfill
\begin{minipage}[c]{0.30\linewidth}
\centering{\scriptsize\color{black!50}\textit{When does it fail?}}\\[-2.5pt]
\begin{tcolorbox}[
  enhanced,
  colback=pathbg,
  colframe=pathframe,
  fonttitle=\bfseries\footnotesize,
  title={Pathologies (\S\ref{sec:diagnosis})},
  coltitle=white,
  colbacktitle=pathframe,
  toptitle=1mm, bottomtitle=1mm,
  boxrule=0.4pt,
  arc=1.5mm,
  top=1mm, bottom=1mm, left=1.5mm, right=1.5mm,
  fontupper=\scriptsize,
  equal height group=overviewpillar,
]
\textbf{Signal Misalignment $\to$ Collapse}\\[2pt]
{\scriptsize\color{black!70}%
\textbullet~Student-Teacher Mismatch\\
\textbullet~Endless Exploration\\
\textbullet~Abrupt Degeneration}
\end{tcolorbox}
\end{minipage}%
\hfill
\begin{minipage}[c]{0.03\linewidth}
\centering
{\color{black!30}\large$\rightarrow$}\\[1pt]
{\color{black!50}\footnotesize\itshape fix}
\end{minipage}%
\hfill
\begin{minipage}[c]{0.30\linewidth}
\centering{\scriptsize\color{black!50}\textit{How to fix it?}}\\[-2.5pt]
\begin{tcolorbox}[
  enhanced,
  colback=regbg,
  colframe=regframe,
  fonttitle=\bfseries\footnotesize,
  title={Regulations (\S\ref{sec:signal-regulation})},
  coltitle=white,
  colbacktitle=regframe,
  toptitle=1mm, bottomtitle=1mm,
  boxrule=0.4pt,
  arc=1.5mm,
  top=1mm, bottom=1mm, left=1.5mm, right=1.5mm,
  fontupper=\scriptsize,
  equal height group=overviewpillar,
]
\textbf{Signal Quality $>$ Teacher Scale}\\[2pt]
{\scriptsize\color{black!70}%
\textbullet~Hard: $\tilde{a}_t{=}\mathrm{clip}(\Delta\ell_t,\,c_{\min},\,c_{\max})$\\
\textbullet~Soft: $\tilde{a}_t{=}\mathrm{sgn}(\Delta\ell_t)\!\cdot\!\log(1{+}|\Delta\ell_t|)$
}
\end{tcolorbox}
\end{minipage}
\end{tcolorbox}
\vspace{-5mm}
\caption{Overview. We characterize OPD's role as an exploration catalyst (\S\ref{sec:role}), diagnose pathologies (\S\ref{sec:diagnosis}), and propose lightweight signal regulations that restore stability without additional compute (\S\ref{sec:signal-regulation}). 
}
\label{fig:overview}
\vspace{-4mm}
\end{figure*}

On-policy distillation (OPD) has rapidly become a standard component of modern LLM post-training pipelines.
By providing token-level supervision from a stronger teacher over the student's own rollouts, OPD has demonstrated remarkable effectiveness~\citep{gu2026minillmonpolicydistillationlarge,qwen3,deepseekai2026deepseekv4}.
Despite its growing adoption, the fundamental training dynamics of OPD remain poorly understood.
Practitioners often observe highly inconsistent behaviors: OPD sometimes improves performance~\cite{lu2025onpolicydistillation}, yet in other cases becomes unstable~\cite{opd-var-1,zhu2026facesonpolicydistillationpitfalls}, collapses exploration, even underperforms outcome-based reinforcement learning~\cite{li2026rethinkingonpolicydistillationlarge}.
This raises our systematic empirical study, we aim to (1) explore what benefits OPD brings ({{Role}}\S~\ref{sec:role}), (2) when OPD fails (Pathologies\S~\ref{sec:diagnosis}), and (3) how to improve it from the previous findings (Regulations\S~\ref{sec:signal-regulation}).

\vspace{0.1mm}
\noindent \textbf{Role}\quad 
OPD simultaneously embodies two distinct features: it is formulated as knowledge distillation~\cite{gu2026minillmonpolicydistillationlarge}, yet operates like an on-policy RL algorithm driven by dense token-level rewards~\cite{gopd}. 
This dual nature motivates us to investigate whether OPD acts as a standard distillation mechanism that expands the student's capability boundary~\cite{hinton2015distillingknowledgeneuralnetwork}, or as an RL framework that merely reshapes trajectories within the base model's existing capability space~\cite{reasoning-bound,RL-no-up}.
Our empirical evaluations strongly support the latter RL-centric view. 
We find that while OPD consistently accelerates early-stage learning, it rarely alters the asymptotic performance ceiling achievable by the base student. 
This acceleration stems from the dense, token-level guidance provided along student-generated trajectories, which effectively helps the student discover correct reasoning paths already reachable within its inherent capability bounds. 
Furthermore, under a fixed computational budget, scaling problem diversity (prompt coverage) yields significantly higher accuracy than scaling per-problem sampling depth. 
Together, these findings demonstrate that OPD functions primarily as an exploration catalyst that reshapes the trajectory distribution, rather than a mechanism for fundamental capability expansion.

\vspace{0.1mm}
\noindent \textbf{Pathologies}\quad
Viewing OPD as an exploration guidance mechanism immediately uncovers a critical dependency: its effectiveness relies entirely on the fidelity of the guidance signal itself. 
While faithful teacher preferences steer the student toward correct trajectories, distorted signals inevitably bias exploration toward degenerate behaviors.
Our analysis reveals two major sources of such guidance corruption, as shown in Figure~\ref{fig:teaser-toy}.
\begin{itemize}[leftmargin=*,nosep]
    \item \textbf{Student-Teacher Mismatch}: When a severe capability gap exists between the teacher and the student, the teacher’s preference allocations on student-generated trajectories decouple from actual quality, generating misleading guidance. 
    This is rooted not in the teacher's absolute strength, but in the mismatch between teacher preferences and the student's rollout distribution.
    \item \textbf{Length Exploitation}: Because token-level signals are aggregated over sequences of varying lengths, the standard optimization objective inadvertently admits degenerate shortcuts. 
    The student exploits this via either  \textit{Endless Exploration} (filler tokens to dilute negative advantages), or  \textit{Abrupt Degeneration} (premature truncation for high aggregated advantage).
\end{itemize}


\vspace{0.1mm}
\noindent \textbf{Regulations}\quad 
We further explore how to eliminate these pathologies.
Intuitive methods include cold-start teacher-student alignment finetuning~\cite{opd-var-1,li2026rethinkingonpolicydistillationlarge}, which are effective but require additional computation.
We investigate two token-level regulation strategies designed to suppress pathological incentives while preserving useful exploration guidance: \emph{Hard Clipping} and \emph{Soft Log-scale Compression}.
Across seven reasoning benchmarks, these interventions consistently stabilize training, eliminate previous pathologies, and substantially outperform naive OPD, RLVR, and prior OPD variants~\cite{eopd}. 
Notably, our regulated framework with a modest teacher decisively surpasses state-of-the-art methods relying on a massive 30B teacher~\cite{gopd,hou2026uni}, demonstrating that the quality of the optimization signal ultimately governs distillation success over brute-force teacher scaling.
Overall, this work makes three contributions:



\begin{itemize}[leftmargin=*,nosep]
    \item We empirically frame OPD as an exploration guidance rather than a capability expander.
    \item We pinpoint guidance corruption, specifically distribution mismatch and length exploitation, as the root cause of OPD failures.
    \item We demonstrate that signal regulations effectively improve OPD stability and effectiveness across diverse reasoning tasks.
\end{itemize}

\section{Preliminaries}

\subsection{On-Policy Distillation (OPD)}\label{sec:opd-prelim}

Let $x$ denote a prompt and $\mathbf{y} = (y_1, \ldots, y_T)$ a response over vocabulary $\mathcal{V}$. We consider a fixed teacher $\pi_\mathrm{T}$ and a trainable student $\pi_\theta$, with $\pi(\mathbf{y}|x) = \prod_{t=1}^T \pi(y_t | \mathbf{y}_{<t}, x)$.

\vspace{1mm}
\noindent \textbf{Objective} \quad
OPD trains the student to minimize the reverse KL divergence from student to teacher:
\vspace{-4pt}
\begin{equation}
\small
  \label{eq:rkl}
  \mathcal{L}_\mathrm{OPD}(\theta)
  = \mathrm{KL}[\pi_\theta \| \pi_\mathrm{T}]
  = -\mathbb{E}_{x,\,\mathbf{y}\sim\pi_\theta}
    \!\left[\log\frac{\pi_\mathrm{T}(\mathbf{y}|x)}{\pi_\theta(\mathbf{y}|x)}\right].
\end{equation}
Because the expectation is taken over the student's own samples, the signal is evaluated on sequences the student actually generates, avoiding the distributional mismatch inherent in off-policy distillation.


\noindent \textbf{Per-token advantage} \quad
Expanding the sequence-level KL token-by-token yields per-token log-ratio:
\vspace{-2pt}
\begin{equation}
\small
  \label{eq:delta_lp}
  \Delta\ell_t \triangleq \log \frac{\pi_{\mathrm{T}}(y_t \mid \mathbf{y}_{<t}, x)}{\pi_\theta(y_t \mid \mathbf{y}_{<t}, x)}, \quad \mathbf{y} \sim \pi_\theta(\cdot \mid x).
\vspace{-1pt}
\end{equation}
Intuitively, $\Delta\ell_t > 0$ indicates the teacher assigns higher probability to token $y_t$ than the student, providing a positive learning signal; $\Delta\ell_t < 0$ indicates the student over-weights $y_t$ relative to the teacher.

\vspace{1mm}
\noindent \textbf{Policy optimization} \quad
OPD is implemented via policy gradient with $\Delta\ell_t$ as the per-token advantage. We apply clipping~\cite{schulman2017ppo} on the importance ratio to stabilize updates:
\vspace{-3pt}
\begin{equation}
  \label{eq:ppo-clip}
  \small
  \mathcal{L}_\mathrm{clip}(\theta) = -\mathbb{E}_t\!\left[\min\!\left(r_t \,\tilde{a}_t,\; \mathrm{clip}(r_t, 1{-}\epsilon, 1{+}\epsilon)\,\tilde{a}_t\right)\right],
\vspace{-3pt}
\end{equation}
where $r_t = \pi_\theta(y_t \mid \mathbf{y}_{<t}, x) \,/\, \pi_{\theta_\mathrm{old}}(y_t \mid \mathbf{y}_{<t}, x)$ is the importance ratio and $\tilde{a}_t$ is a regulated form of $\Delta\ell_t$. This ratio clipping constrains how far the policy moves per update but does not alter the reward landscape defined by the teacher signal.

\subsection{Empirical Framework and Setup}
In this section, we establish the experiment bases for our study. 
To ensure a systematic investigation and the reproducibility of our findings, we strictly adhere to the experimental configurations described below across subsequent training and evaluation.

\paragraph{Training Configuration}

Our empirical analysis uses Qwen3-1.7B-Base as the student and Qwen3-series post-trained models of varying scales as teachers. To isolate true distillation gains from mere thinking-mode shifts, we enforce a strict non-thinking constraint by prefixing all rollouts with a \textit{} block. This simultaneously ensures an aligned prompt format between models~\cite{li2026rethinkingonpolicydistillationlarge}. Trained on the Nemotron-Cascade Math dataset~\citep{nemotron}, we compare models optimized via GRPO~\citep{grpo, dapo} and OPD~\citep{gu2026minillmonpolicydistillationlarge}.



\paragraph{Evaluation}
To provide a robust and statistically sound assessment, we employ the \textbf{avg@32} metric for AMC23~\cite{MAA_AMC}, AIME 2024--2026~\cite{MAA_AIME}, and HMMT25 Feb~\cite{hmmt}, while reporting standard \textbf{pass@1} for MATH500~\cite{hendrycks2021math} and Minerva~\cite{lewkowycz2022minerva} following previous works~\cite{eopd}.

\begin{figure}[t]
    \centering
    \vspace{-2pt}
    \includegraphics[width=\linewidth]{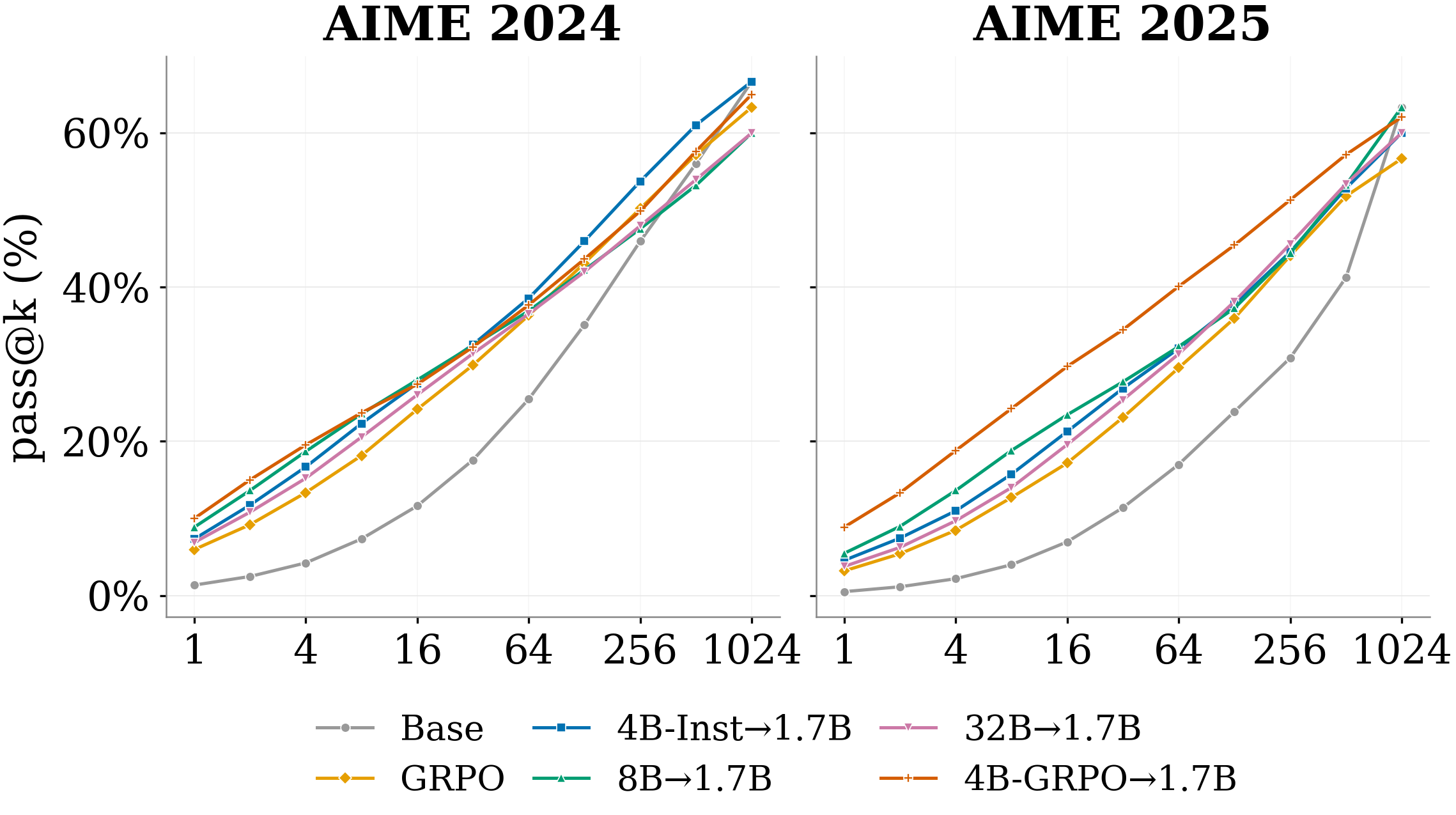}
    \vspace{-8pt}
    \caption{Pass@$k$ performance of Qwen3-1.7B-Base trained via different methods. OPD variants outperform Base and GRPO models in low-$k$. However, the convergence at higher $k$ suggests {OPD accelerates exploration rather than raising the model's performance ceiling}. 
    }
    \label{fig:explore-ceiling}
    \vspace{-3mm}
\end{figure}



\section{Exploring the Role of OPD}\label{sec:role}
Conceptually, OPD operates at the intersection of knowledge distillation~\cite{hinton2015distillingknowledgeneuralnetwork,gu2026minillmonpolicydistillationlarge}, and on-policy RL with dense token-level rewards~\cite{gopd}. 
While standard distillation intuition suggests that the teacher transfers superior intelligence to expand the student's capability boundary, RL primarily reshapes trajectories within the latent capability space already encoded in the base model~\cite{reasoning-bound,RL-no-up}.  
Under large sampling budgets ($k$), the performance of RL-trained models often converges to that of their unaligned base counterparts via brute-force test-time sampling \citep{RL-no-up}. 
This capability convergence motivates a fundamental question for OPD: \textit{Does OPD fundamentally elevate the inherent capability ceiling of the student, akin to distillation, or does it accelerate early-stage exploration efficiency, much like RL?}

To disentangle these two possibilities, we systematically analyze the performance convergence trajectories of student models trained via OPD and RLVR across varying teacher configurations and evaluation sampling size $k$.
We set the Qwen3-1.7B-Base as the student, and Qwen3-4B-instruct-2507, Qwen3-\{4B,8B,32B\} and GRPO trained Qwen3-4B as the teachers.
Then we test the pass@1024 of tuned models to explore their performance ceilings. 

\paragraph{Efficient Exploration rather than Ceiling Breaker}
As shown in Figure~\ref{fig:explore-ceiling}, OPD variants provide a distinct boost in the low-sample regime ($k \leq 64$). 
However, in the high-sample regime, this performance gap narrows. 
The base model eventually converges to the level of the GRPO and OPD-trained models, indicating that OPD, just like RLVR, does not fundamentally inject entirely new capabilities~\cite{RL-no-up}. 
Instead, it acts as an exploration catalyst that re-weights the student's inherent generative space, ensuring optimal solutions are surfaced within the first few attempts.

This characterization highlights a critical vulnerability: 
If the teacher's guidance is misaligned or the objective function contains structural flaws, this catalytic effect will instead accelerate the student's convergence toward degenerate or sub-optimal solutions. 
We dissect these specific pathologies in the following section.

\vspace{-0.1mm}
\noindent \textbf{Prompt Diversity v.s. Rollout Sizes}\quad
Our previous analysis establishes that OPD benefits the student primarily through high-density guidance across diverse problem spaces. This motivates a critical resource-allocation question: {Under a fixed computational budget, should we prioritize \emph{problem coverage} (more distinct prompts) or \emph{rollout depth} (more rollouts per prompt)}? 

\begin{figure}[t]
      \centering
      \includegraphics[width=0.88\linewidth]{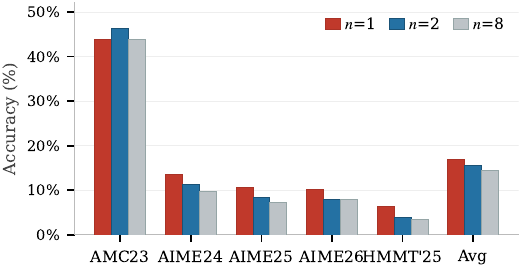}
      \vspace{-6pt}
      \caption{\textbf{Avg@32 across varying prompt-to-rollout configurations.} Under a fixed batch $B$, we vary the rollout count $n$ per
   prompt. {Maximizing problem diversity consistently outperforms a larger sampling size}.}
      \label{fig:roll-n}
      \vspace{-3mm}
  \end{figure}

To investigate this, we evaluate $n{=}1$ ($B$ distinct prompts with $1$ rollout each) against $n{=}2$ and $n{=}8$ configurations under identical total compute. As illustrated in Figure~\ref{fig:roll-n}, the $n{=}1$ setup consistently achieves superior final accuracy. This confirms that OPD's exploration benefits scale primarily with problem diversity rather than per-problem sampling depth. Consequently, we adopt the $n=1$ configuration across all subsequent experiments to maximize distillation efficiency.



\vspace{-2pt}
\begin{tcolorbox}[colback=gray!5, colframe=gray!60, boxrule=0.4pt, left=4pt, right=4pt, top=2pt, bottom=2pt]
\textbf{Finding 1.}
OPD incentivizes the student's exploration efficiency, not performance ceilings.
\end{tcolorbox}

\begin{figure*}[t]
    \centering
    \includegraphics[width=0.83\linewidth]{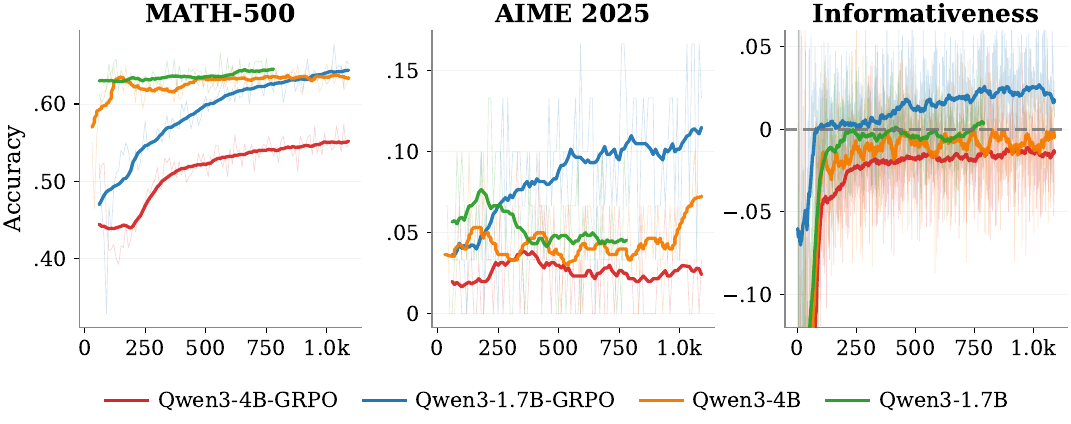}
  \caption{\textbf{Weak teacher outperforms strong teacher in OPD w/ Clip.} Benchmark accuracy (first two panels)   
  and informativeness $\mathcal{I}$ on the validation set for OPD w/ Clip runs distilling from four teachers with varying capabilities into the same {Qwen3-1.7B-Base} student.}
    \label{fig:train-dyna}
    \vspace{-4mm}
\end{figure*}



\section{Pathologies of On-Policy Distillation}
\label{sec:diagnosis}
Given that OPD operates by re-weighting the student’s exploration landscape, its efficacy hinges on the \textit{fidelity} of the reward signal. 
We identify two distinct failure modes where this guidance mechanism breaks down, leading to the stagnation or collapse of student performance. 

First, a \textbf{Student-Teacher Mismatch} (\S\ref{sec:strong-teacher-trap}): we challenge the stronger-is-better dogma by showing that a significant distributional gap between teacher and student can turn the distillation signal into counter-productive noise. 
Second, a \textbf{Length Exploitation} (\S\ref{sec:length-hacking}): the cumulative nature of token-level advantages creates length-dependent shortcuts that allow the student to leverage the objective without genuine reasoning improvements.


\subsection{Student-Teacher Mismatch}
\label{sec:strong-teacher-trap}
A common intuition in knowledge distillation is that a more capable teacher yields a superior student. However, our empirical results challenge this assumption, revealing a complex trade-off between standalone capability and distributional alignment.
\paragraph{Setup}
We continue to employ Qwen3-1.7B-Base as the student. 
To evaluate the student's sensitivity to varying teacher expertise, we select teachers across a controlled spectrum of capabilities (Table~\ref{tab:main}): Qwen3-\{1.7B, 4B\}-GRPO, and Qwen3-\{1.7B, 4B\}. The capability order of teachers is: Qwen3-4B-GRPO $>$ Qwen3-1.7B-GRPO $>$ Qwen3-4B $>$ Qwen3-1.7B.

\paragraph{Early Efficiency is Deceptive}
As shown in the MATH-500 results (Figure~\ref{fig:train-dyna}), weaker teachers such as Qwen3-1.7B and Qwen3-4B lead to rapid accuracy gains during the initial training steps. 
This high initial learning efficiency suggests that teachers with lower performance are more learnable for a base student.
However, this early speed is deceptive. 
These models quickly hit a performance plateau, particularly on challenging benchmarks, due to their capability boundaries. 
This observation implies that initial learning velocity is an unreliable proxy for the final capability ceiling.

\begin{figure*}[t]
    \centering
    \includegraphics[width=0.9\linewidth]{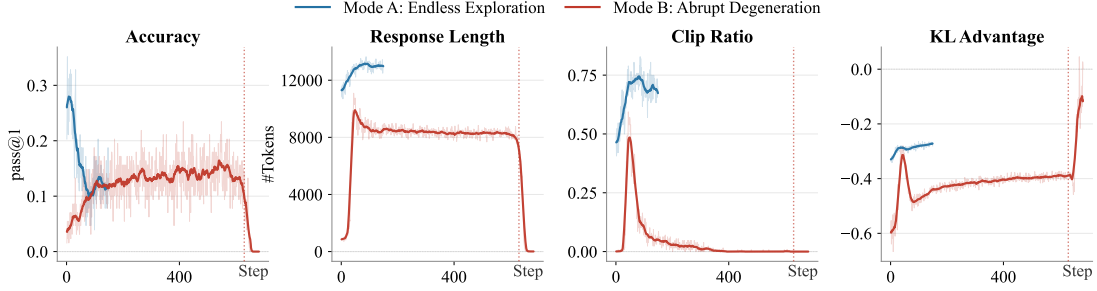}
    \vspace{-8pt}
    \caption{Length exploitation illustration. Advantage increases, while accuracy drops. Teacher$\rightarrow$Student: Qwen3-4B-GRPO$\rightarrow$Qwen3-1.7B (Mode A), and Qwen3-1.7B-GRPO$\rightarrow$Qwen3-1.7B-Base (Mode B). }
    \label{fig:failure_modes}
\end{figure*}

\paragraph{The Failure of the Strongest Teacher}
The most striking observation is the failure of the most capable model in our suite: Qwen3-4B-GRPO. Despite having the highest performance, it is paradoxically the least effective teacher. As shown across both benchmarks in Figure~\ref{fig:train-dyna}, the student distilled from 4B-GRPO remains nearly stagnant, with its accuracy on AIME 2025 hovering near 2\% throughout the training process.

This inversion is a significant anomaly: a 4B RL-trained model provides significantly worse guidance than its 1.7B counterpart. 
It suggests that standalone performance does not translate into teaching quality.
Ultimately, 1.7B-GRPO yields the highest final accuracy, outperforming teachers that are either easier to learn from or smarter in isolation (4B-GRPO). 
This indicates that the most effective teacher is not necessarily the one who is easiest to mimic or the one with the strongest performance, but the one whose reasoning paths are both advanced and bridgeable for the student.

\paragraph{Diagnosis}
To explain this paradox, we introduce the \textbf{Informativeness} metric $\mathcal{I}$, which measures whether the teacher's token-level signal discriminates correct from incorrect student rollouts: 
\begin{equation}
\small
\mathcal{I} = \mathbb{E}_{\mathbf{y} \sim \pi_\theta}\!\bigl[\,\overline{\Delta \ell} \mid r = 1\bigr] - \mathbb{E}_{\mathbf{y} \sim \pi_\theta}\!\bigl[\,\overline{\Delta \ell} \mid r = 0\bigr],
\end{equation}
where $\overline{\Delta \ell} = \frac{1}{|\mathbf{y}|}\sum_t(\log \pi_T(y_t) - \log \pi_\theta(y_t))$ is the mean per-token log-ratio, $r$ denoting the rollout accuracy.
A positive $\mathcal{I}$ indicates that the teacher assigns a higher probability to tokens in correct rollouts than incorrect ones, providing an outcome accuracy aligned optimization.
The rightmost panel of Figure~\ref{fig:train-dyna} reveals that informativeness ($\mathcal{I}$) precisely predicts distillation trajectories, clustering into three distinct dynamic modes:
\textbf{(1) Passable ($\mathcal{I} \approx 0$):} Qwen3-1.7B and 4B drive rapid early gains on MATH500 but quickly plateau on the harder AIME 2025; \textbf{(2) Misleading ($\mathcal{I} < 0$):} Qwen3-4B-GRPO provides persistently anti-correlated signals, leading to early accuracy drops and a low ceiling; and \textbf{(3) Recovery}: Qwen3-1.7B-GRPO starts negative but recovers past zero, unlocking a late-stage surge that ultimately outperforms all others.



\begin{tcolorbox}[colback=gray!5, colframe=gray!60, boxrule=0.4pt, left=4pt, right=4pt, top=2pt, bottom=2pt]
\textbf{Finding 2.} 
The effectiveness of OPD depends on whether the teacher can give rewards aligned with Informativeness~$\mathcal{I}$ on student's rollouts, rather than the teacher's absolute capability.
\end{tcolorbox}

\subsection{Length Exploitation}
\label{sec:length-hacking}
\begin{figure}
    \centering
    \includegraphics[width=\linewidth]{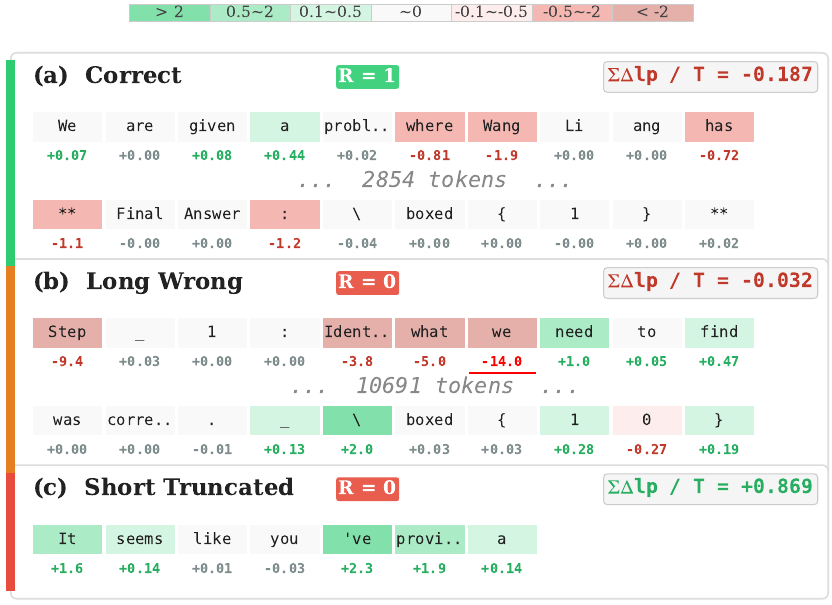}
    \vspace{-5pt}
    \caption{\textbf{(a)}~A correct rollout receives a mildly negative token-mean.           
      \textbf{(b)}~\emph{Mode~A (Endless Exploration)}: a long wrong rollout with negative scores diluted across tokens, failing to penalize the incorrect response.
      \textbf{(c)}~\emph{Mode~B (Abrupt Degeneration)}: a short, fully teacher-preferred rollout prefix that yields a strongly positive token-mean despite an incorrect outcome.
      }
      \vspace{-3mm}
    \label{fig:token-adv}
\end{figure}

While the previous section highlighted how external teacher selection destabilizes training, we now turn to an intrinsic vulnerability embedded within the standard OPD objective itself. The sequence-level guidance is propagated via the token-level distillation advantage, $a_t = \log\pi_T(y_t|y_{<t}) - \log\pi_\theta(y_t|y_{<t})$. Unlike outcome-based rewards that are inherently bounded and strictly tied to final correctness, $a_t$ is unconstrained, exhibits high variance, and is fundamentally decoupled from the actual accuracy of the response.
Since the token-mean advantage $\bar{a} = \frac{1}{T}\sum_t a_t$ depends on both the per-token signal and the response length $T$, the student can exploit this interaction to inflate its advantage without improving accuracy. 
We identify two complementary failure modes, illustrated in Figure~\ref{fig:failure_modes} and Figure~\ref{fig:token-adv}.

\paragraph{Endless Exploration (Mode A)}
When a rollout yields an incorrect reasoning path, it incurs a severe overall negative advantage. Crucially, because standard OPD objectives optimize the sequence-averaged advantage $\bar{a} = \frac{1}{T}\sum_{t=1}^{T} a_t$, the student discovers a degenerate shortcut to evade this penalty: appending low-information filler tokens to inflate the denominator $T$. 
Formally, suppose the core reasoning block terminates at step $T_0$ with a cumulative negative signal $A_{\text{core}} = \sum_{t=1}^{T_0} a_t < 0$. By continually generating redundant tokens up to a prolonged length $T$ ($T \gg T_0$), the aggregated advantage decomposes as:
\begin{equation}
\label{eq:long_hack}
\small
\bar{a}
\;=\; \frac{A_{\text{core}}}{T} + \frac{T - T_0}{T}\,\bar{a}_{\text{filler}}.
\end{equation}
Eq.~\eqref{eq:long_hack} reveals a severe structural pathology: 
By scaling the sequence length $T$, the student can asymptotically suppress the magnitude of its negative advantage ($\lim_{T \to \infty} \bar{a} = 0$), effectively washing out the penalty gradient ($\nabla_\theta \mathcal{L} \approx 0$). As a result, the student is never penalized for its logical failures, trapping the training loop in an unproductive state of endless, verbose exploration without any actual improvement in reasoning accuracy.

As shown in Figure~\ref{fig:failure_modes}, both runs exhibit a common pattern: response lengths grow sharply.
Once truncated, the model fails to produce valid final answers, causing a sharp collapse in accuracy.
This pathology is particularly severe in the {Qwen3-4B}$\rightarrow${Qwen3-1.7B} setting, which response length grows until hitting the predefined maximum limit.   
Crucially, despite the drop in performance, the aggregated advantage continues to climb, demonstrating a clear pattern of advantage hacking.


\begin{table*}[t]
      \centering
      \setlength{\tabcolsep}{5pt}
      \renewcommand{\arraystretch}{1.15}
      \resizebox{\textwidth}{!}{%
      \begin{tabular}{lll cc ccccc c}
      \toprule
      \multirow{2}{*}{\textbf{Teacher}} & \multirow{2}{*}{\textbf{Student}}
        & \multirow{2}{*}{\textbf{Method}}
        & \multicolumn{2}{c}{\textit{pass@1}}
        & \multicolumn{5}{c}{\textit{avg@32}} 
        & \multirow{2}{*}{\textbf{Avg}} \\
      \cmidrule(lr){4-5}\cmidrule(lr){6-10}
      & & & \textbf{MATH500} & \textbf{Minerva}
            & \textbf{AMC'23} & \textbf{AIME'24} & \textbf{AIME'25} & \textbf{AIME'26} & \textbf{HMMT'25} & \\
      \midrule
      \multirow{4}{*}{---}
        & Qwen3-1.7B-Base    & --   & 18.0 & 4.1   & 6.3   & 1.7 & 0.4 & 0.6 & 0.1  & 4.5 \\
        & Qwen3-1.7B         & --   & 73.0 & 32.7   & 42.5   & 13.4 & 9.8  & 7.8   & 3.7   & 26.1 \\
        & Qwen3-8B           & --   & 83.4 & 47.8   & 66.9   & 29.1 & 20.9 & 14.8   & 10.4   & 39.0 \\
        & Qwen3-4B-Inst-2507 & --   & 93.6  & 52.9  & 94.2   & 59.3 & 47.4 & 55.8   & 31.4 & 62.1 \\

      \hdashline
      ---   & Qwen3-1.7B-Base    & GRPO & 63.1 & 29.5 & 43.0 &  9.2 &  4.9 & 4.7   & 2.3   & 22.4 \\
      ---   & Qwen3-1.7B         & GRPO & 90.0    & 45.9   & 81.7   & 41.9 & 32.5 & 37.5   & 21.7   & 50.2 \\
      ---   & Qwen3-4B           & GRPO & 93.6     & 55.5   & 92.9   & 64.4 & 57.1 & 57.3   & 37.3   & 65.4 \\

      \midrule
      \rowcolor{gray!6} \multicolumn{11}{l}{\textit{\textbf{Teacher: Qwen3-8B}}} \\
        & Qwen3-1.7B-Base & KD   & 63.8 & 30.2 & 40.6 &  9.2 &  4.6 & -- & -- & -- \\
       & Qwen3-1.7B-Base & EOPD & 68.7 & 30.2 & 41.9 & 10.4 &  5.8 & -- & -- & -- \\
      
      \midrule
      \rowcolor{gray!6} \multicolumn{11}{l}{\textit{\textbf{Teacher: Qwen3-1.7B-GRPO}}} \\
      & Qwen3-1.7B-Base & OPD      & 75.6 & \underline{36.3} & 43.9 & \underline{13.6} & 10.6 & 10.3 & \underline{6.3} & 28.1 \\
      & Qwen3-1.7B-Base & \ +Clip  & \underline{77.0} & 35.4 & \underline{45.7} & 12.1 & \underline{13.7} & \underline{10.5} & {5.8} & \underline{28.6} \\
      & Qwen3-1.7B-Base & \ +Scale & \textbf{79.3} & \textbf{40.1} & \textbf{46.3} & \textbf{13.8} & \textbf{15.6} & \textbf{10.6} & \textbf{7.3} & \textbf{30.4} \\
      
      \midrule

      \rowcolor{gray!6} \multicolumn{11}{l}{\textit{\textbf{Teacher: Qwen3-4B-GRPO}}} \\
      & Qwen3-1.7B-Base & OPD      & \underline{74.4} & \textbf{34.2} & \underline{44.3} & \underline{11.6} & 9.4 & \underline{8.6} & 4.7 & {26.7} \\
      & Qwen3-1.7B-Base & \ +Clip  & \textbf{76.2} & \underline{33.8} & \textbf{47.4} & 10.8 & \textbf{13.5} & {\textbf{8.7}} & \textbf{6.5} & \textbf{28.1} \\
      & Qwen3-1.7B-Base & \ +Scale & {74.3} & {\textbf{34.2}} & 41.2 & \textbf{12.8} & \underline{12.1} & 7.8 & \underline{5.4} & \underline{26.8} \\
      \bottomrule
      \end{tabular}%
      }
      \caption{Performance of different methods on benchmarks. Among all evaluated downstream student models (bottom two blocks), the overall best scores are \textbf{bolded} and the second best are \underline{underlined}.}
      \label{tab:main}
\end{table*}

\paragraph{Short-Sequence Exploitation (Mode B)}
In this situation, the student learns to truncate responses prematurely.
By generating a high-confidence preamble (e.g., ``We are given the problem...'') and immediately emitting \texttt{EOS}, the model captures the favorable prefix advantage $\sum_{t=1}^{T_{\text{pre}}} a_t$ while avoiding the riskier reasoning steps where $a_t$ may turn negative.
Formally, whenever the prefix advantage exceeds the full-sequence advantage: 
\begin{equation}
\label{eq:short_hack}
\small
\frac{1}{T_{\text{pre}}}\sum_{t=1}^{T_{\text{pre}}} a_t \;>\; \frac{1}{T}\sum_{t=1}^{T} a_t,
\end{equation}
The optimizer is incentivized to favor shorter trajectories regardless of correctness.
As shown in Figure~\ref{fig:failure_modes}, in the late training stage, the student only output extremely short and safe responses, but receives a high reverse KL advantage from the teacher. 
Once the policy locks into short outputs, the gradient reinforces this equilibrium.

\paragraph{Dignosis}
Both modes exploit the same structural flaw: the token-level advantage $a_t$ interacts with sequence length in a way that creates \emph{gradients unrelated to reasoning correctness}.
Mode~A exploits unfavorable sequences by padding late; Mode~B exploits favorable prefixes by cutting early.
Together they form a squeeze: the optimizer can always improve the objective by manipulating length rather than improving reasoning, regardless of whether the current rollout outcome is positive or negative.

\begin{tcolorbox}[colback=gray!5, colframe=gray!60, boxrule=0.4pt, left=4pt, right=4pt, top=2pt, bottom=2pt]
\textbf{Finding 3.}
The unstable token-level objective introduces degenerate optimization shortcuts at both length extremes.
\end{tcolorbox}

\section{Regulations}\label{sec:signal-regulation}

Prior work stabilizes OPD through mechanisms such as utilizing off-policy data~\cite{opd-var-1}, or scoring top-$k$ tokens beyond the predicted one to smooth gradient estimates~\cite{deepseekai2026deepseekv4}.
These approaches are effective but incur substantial overhead, maintaining full vocab replay buffers during training~\cite{deepseekai2026deepseekv4} or warm-starting students with SFT on teacher rollouts for cold start~\cite{opd-var-1}.
We instead pursue modifications that stabilize OPD \emph{without additional compute}, operating entirely within the existing training loop and motivated by the diagnostic findings of the previous section.


\begin{figure}                                                    \centering                         
      \includegraphics[width=0.95\linewidth]{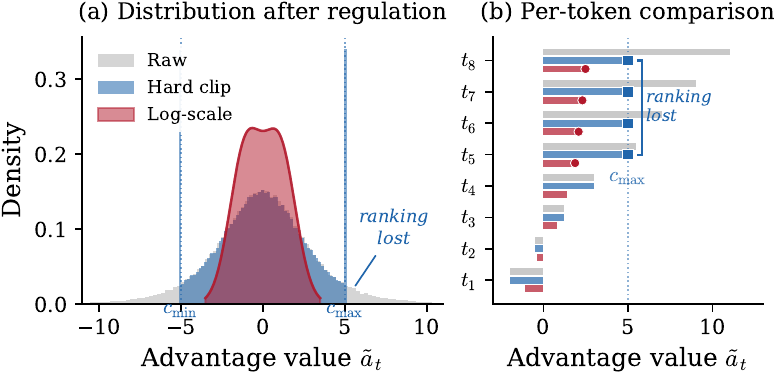}                                   \caption{Hard Clipping collapses all tokens beyond the $[c_{\min}, c_{\max}]$ to the same value (ranking lost). Log-scale compression preserves ordering across the full range.}    
      \label{fig:signal-regulation}   
      \vspace{-3mm}
\end{figure}

\subsection{Stabilizing OPD with Signal Regulation}        
\label{sec:regulation}

Motivated by the aforementioned pathologies, we transition from diagnosing length exploitation to developing proactive in-loop remedies. 
To prevent extreme token-level log-ratios from dominating the policy gradient, which \textbf{exacerbates the teacher–student mismatch and enables the student to exploit length through extreme signals}, we introduce token-level signal regulation.

Our design follows two principles: \textbf{(1) Preference preservation}, regulation must not reverse the teacher's ordinal signal; \textbf{(2) Outlier suppression}, pathologically large magnitudes should be dampened without discarding fine-grained token-level information.
We investigate two complementary paradigms, representing \emph{Hard truncation} and \emph{Soft compression} respectively (Figure~\ref{fig:signal-regulation}):


\begin{itemize}[leftmargin=*,nosep]
    \item \textbf{Hard Clipping.} As a representative of rigid thresholding, this method truncates the advantage into a deterministic bound to eliminate severe outliers while preserving signal diversity:
    \begin{equation}
    \small
    \label{eq:adv-clip}
    \tilde{\Delta\ell}_t = \mathrm{clip}\!\bigl(\Delta\ell_t,\; c_{\min},\; c_{\max}\bigr).
    \end{equation}

    \item \textbf{Soft Log-Scale Compression.} As a smooth alternative, this method dampens unbounded signals without discarding the relative rankings beyond fixed bounds:
    \begin{equation}
    \label{eq:adv-logscale}
    \small
    \tilde{\Delta\ell}_t = \mathrm{sign}(\Delta\ell_t) \cdot \log\!\bigl(1 + |\Delta\ell_t|\bigr).
    \end{equation}
    The choice of $\log(1+|x|)$ is motivated by its dual-regime properties. Near zero, it approximates a linear mapping ($\tilde{\Delta\ell}_t \approx \Delta\ell_t$), fully preserving the teacher's fine-grained preferences. Crucially, when $|\Delta\ell_t| \to \infty$, its sub-linear growth radically compresses the magnitude of unbounded values. 
\end{itemize}

\paragraph{Experiment Setup} 
To explore the performance of the above two optimization techniques for OPD, we utilize Qwen3-1.7B-GRPO, Qwen3-4B-GRPO as the teachers, and Qwen3-1.7B-Base and Qwen3-1.7B as the students. 
For better comparison, we select current OPD variants as counterparts~\cite{eopd,gopd,hou2026uni}.
The performance of Qwen3-1.7B-Base as the student is shown in Table~\ref{tab:main}, and the scores of Qwen3-1.7B are shown in Table~\ref{tab:unified_math_results}.

\begin{figure}
    \centering
    \includegraphics[width=\linewidth]{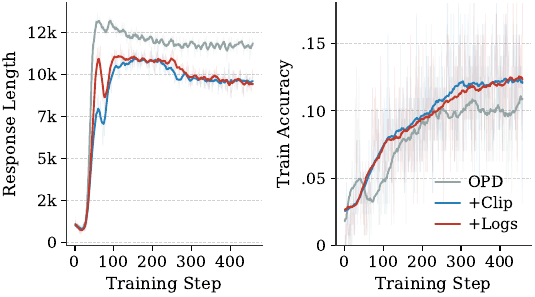}
    \caption{Training dynamics of OPD variants. Teacher: Qwen3-4B-GRPO; Student: Qwen3-1.7B-Base. Vanilla OPD suffers greater 
  length exploitation, whereas Clip produces \textbf{shorter responses and achieves higher accuracy} than naive OPD.}
    \label{fig:dya-from-4b}
    \vspace{-3mm}
\end{figure}

\subsection{Analysis}

\paragraph{Signal Regulation is Effective.}
As shown in Table~\ref{tab:main}, both regulation variants yield clear improvements over naive OPD.
With the 1.7B-GRPO teacher, OPD+Clip brings substantial gains across all benchmarks.
Comparing the two variants, log-scale compression outperforms hard clipping for Qwen3-1.7B-Base, likely because it retains the teacher's full token ranking rather than flattening extreme values to the clip boundary.

\noindent \textbf{Alleviating the Student-Teacher Mismatch.}\quad
Our experiments further demonstrate that {naively scaling up the teacher does not guarantee better student performance}. 
Hard clipping partially rescues this, lifting the 4B-guided student back to 28.1 of average score. 
Interestingly, log-scale compression, which excelled with the 1.7B teacher, fails to yield comparable gains here, as the training dynamics shown in Figure~\ref{fig:dya-from-4b}. 
We attribute this to the nature of the distributional shift: a much stronger teacher heavily mixes genuine correctness cues with biases in its advantages. 
Log-scale compression preserves this noise, whereas hard clipping aggressively truncates it. 
This yields a crucial insight: \textbf{when the teacher-student gap is large, aggressively truncating extreme signals is more effective than compressing them.}

\paragraph{Impact of Student Capacity}

Comparing Tables~\ref{tab:main} and~\ref{tab:unified_math_results} reveals a striking impact between student capacity and teacher strength. 
When the student is upgraded from a base model to the post-trained {Qwen3-1.7B}, the initial distributional mismatch is significantly mitigated. 
{Consequently, the performance hierarchy flips:} {Qwen3-4B-GRPO} teacher now delivers the best results, decisively outperforming the {Qwen3-1.7B-GRPO} teacher. 
This shift confirms that the trap stems from the capability gap with student-teacher mismatch. 
{Crucially, by resolving this mismatch,} our regulated OPD with a 4B teacher ultimately surpasses previous methods distilled from a larger 30B teacher. 
This demonstrates that proper capacity matching combined with signal regulation can eclipse brute-force parameter scaling.

\section{Related Work}

\paragraph{On-Policy Distillation.}
\begin{table}[t]
    \small
    \centering
    \setlength{\tabcolsep}{5pt} 
    \renewcommand{\arraystretch}{1.2}
    \begin{tabular}{l cccc}
    \toprule
    \textbf{Method} & \textbf{AIME24} & \textbf{AIME25} & \textbf{AIME26} & \textbf{HMMT25} \\
    \midrule
    
    \rowcolor{gray!10} \multicolumn{5}{c}{\textit{\textbf{Teacher: Qwen3-30B-A3B-Instruct-2507}}} \\
    \rowcolor{gray!10} Teacher     & 74.7 & 62.8 & 63.3 & 44.2 \\
    \textbf{UniOPD}      & 35.2 & 30.7 & --   & 17.7 \\
    \textbf{GOPD}       & 37.3 & 31.5 & --   & 16.2 \\
    
    \midrule 
    \rowcolor{gray!10} \multicolumn{5}{c}{\textit{\textbf{Teacher: Qwen3-1.7B-GRPO}}} \\
    \rowcolor{gray!10} Teacher     & 41.9 & 32.5 & 37.5 & 21.7 \\
    \textbf{CLIP}        & \underline{44.4} & \underline{32.5} & \underline{35.9} & \textbf{20.9} \\
    
    \midrule
    \rowcolor{gray!10} \multicolumn{5}{c}{\textit{\textbf{Teacher: Qwen3-4B-GRPO}}} \\
    \rowcolor{gray!10} Teacher     & 64.4 & 57.1 & 57.3 & 37.3 \\
    \textbf{CLIP}        & \textbf{45.2} & \textbf{37.5} & \textbf{36.0} & \underline{19.0} \\
    \bottomrule
    \end{tabular}
    \caption{Performance of Qwen3-1.7B student tuned by OPD+Clip (CLIP) and previous methods~\cite{hou2026uni,gopd}, which further proves that \textit{a stronger teacher cannot ensure better student performance}. The \textbf{best} and the \underline{second best} scores are marked.  }
    \label{tab:unified_math_results}
    \vspace{-3mm}
\end{table}

Early formulations of OPD optimized a reverse-KL objective to restrict the student to the teacher's high-probability modes~\citep{gu2026minillmonpolicydistillationlarge,GKD2024}. Subsequent work reframed the teacher's per-token log-ratios as dense rewards capable of shaping exploration trajectories~\citep{gopd}, fueling its rapid adoption in scale post-training~\citep{deepseekai2026deepseekv4,qwen3,glm5team2026glm5vibecodingagentic,Nemotron-Cascade-2}.
Despite this widespread use, the failure conditions of this dense guidance remain poorly understood. 
Prior exploratory studies attribute training instability to factors like thinking-pattern compatibility, offering external fixes such as prompt alignment or cold-start initialization~\cite{li2026rethinkingonpolicydistillationlarge,opd-var-1,yu2026dopddualonpolicydistillation}. 
Our work takes a fundamentally parallel yet deeper trajectory: we diagnose signal-level pathologies (student-teacher mismatch and length exploitation) and propose zero-overhead in-loop regulations (clipping and log-scale compression) that require neither off-policy data nor prompt filtering.



\paragraph{Capacity Gap in Knowledge Distillation.}

A recurring phenomenon in knowledge distillation (KD) is that excessively capable teachers can unexpectedly degrade student performance. Specifically, \citet{cho2019efficacy} noted that higher benchmark accuracy in teachers does not inherently translate into superior distillation efficacy.  To mitigate the structural mismatches between models, various strategies have been proposed: \citet{kd-from-strong-1} identified a severe prediction discrepancy between teachers and students, proposing the utilization of the teacher's recommendation label preference order to alleviate the gap, while \citet{mirzadeh2020improved} introduced intermediate teacher assistants to bridge this capacity rift.
Within the context of reasoning, this learnability barrier remains equally pronounced. For instance, \citet{li2025learnability} observed that complex, long reasoning traces generated by strong teachers consistently underperform simpler, teacher-free approaches when distilled into smaller student models. \citet{li2026rolereasoningpatternsgeneralization} demonstrated that the alignment of thinking patterns between teachers and students profoundly impacts distillation outcomes.

Our work extends this classic line of inquiry into the on-policy regime.  Rather than merely documenting the empirical decay caused by strong teachers, we offer a precise, signal-level explanation for \emph{why} and \emph{when} these advanced guidance signals corrupt on student-generated rollouts. Furthermore, we demonstrate that the capability gap can be mitigated via lightweight, in-loop signal regulation.

\section{Conclusion}
\label{sec:conclusion}

In this work, we look inside OPD to shift the paradigm from empirical scaling to rigorous signal-level diagnosis. 
We first established that OPD acts as an \emph{exploration catalyst} which benefits more from problem diversity.
Then we systematically expose two critical pathologies: \emph{length exploitation} via redundant token padding, and the \emph{Student-Teacher Mismatch}, where severe capabilities mismatch causes negative impacts. 
To restore guidance fidelity, we introduce lightweight signal regulations that require zero off-policy overhead. Our empirical results demonstrate that regulating token-level advantage dynamics effectively eliminates these failures, enabling a 4B teacher to surpass configurations distilled from larger models,  which is \textit{signal quality, not teacher scale, governs OPD success}.


\section*{Limitations}
Although our proposed hard clipping and soft log-scale compression strategies effectively restore guidance fidelity and stabilize training, they introduce hyperparameter heuristics that require empirical tuning based on the specific teacher-student capacity gap. A dynamic, adaptive regulation framework could be further explored.
Our systematic empirical study primarily focuses on mathematical reasoning benchmarks, where outcomes are easily verifiable and reasoning chains are dense. 
The behavior of regulated OPD in open-ended text generation or knowledge-intensive QA tasks has not been fully verified.

\bibliography{custom}

\begin{thebibliography}{31}
\providecommand{\natexlab}[1]{#1}

\bibitem[{Agarwal et~al.(2024)Agarwal, Vieillard, Stanczyk, Ramos, Geist, and Bachem}]{GKD2024}
Rishabh Agarwal, Nino Vieillard, Piotr Stanczyk, Sabela Ramos, Matthieu Geist, and Olivier Bachem. 2024.
\newblock \href {https://openreview.net/forum?id=mWRngkvIki} {{GKD}: Generalized knowledge distillation for auto-regressive sequence models}.
\newblock In \emph{Proceedings of the 12th International Conference on Learning Representations ({ICLR})}.

\bibitem[{Chen et~al.(2024)Chen, Qin, WANG, Zhou, and Che}]{reasoning-bound}
Qiguang Chen, Libo Qin, Jiaqi WANG, Jingxuan Zhou, and Wanxiang Che. 2024.
\newblock \href {https://openreview.net/forum?id=pC44UMwy2v} {Unlocking the capabilities of thought: A reasoning boundary framework to quantify and optimize chain-of-thought}.
\newblock In \emph{The Thirty-eighth Annual Conference on Neural Information Processing Systems}.

\bibitem[{Cho and Hariharan(2019)}]{cho2019efficacy}
Jang~Hyun Cho and Bharath Hariharan. 2019.
\newblock \href {https://arxiv.org/abs/1910.01348} {On the efficacy of knowledge distillation}.
\newblock In \emph{Proceedings of the IEEE/CVF International Conference on Computer Vision (ICCV)}, pages 4794--4802.

\bibitem[{DeepSeek-AI(2026)}]{deepseekai2026deepseekv4}
DeepSeek-AI. 2026.
\newblock Deepseek-v4: Towards highly efficient million-token context intelligence.

\bibitem[{Dekoninck et~al.(2026)Dekoninck, Jovanović, Gehrunger, Rögnvaldsson, Petrov, Sun, and Vechev}]{hmmt}
Jasper Dekoninck, Nikola Jovanović, Tim Gehrunger, Kári Rögnvaldsson, Ivo Petrov, Chenhao Sun, and Martin Vechev. 2026.
\newblock \href {https://arxiv.org/abs/2605.00674} {Beyond benchmarks: Matharena as an evaluation platform for mathematics with llms}.

\bibitem[{GLM-5-Team et~al.(2026)GLM-5-Team, Zeng, Lv, Hou, Du, Zheng, Chen, Yin, Ge, Huang, Xie, Zhu, Yin, Wang, Pan, Zeng, Zhang, Wang, Chen, Zhang, Jiao, Guo, Wang, Du, Wu, Wang, Li, Fan, Zhong, Liu, Zhao, Du, Dong, Lu, Shuang-Li, Cao, Liu, Jiang, Chen, Zhang, Huang, Dong, Xu, Wei, An, Niu, Zhu, Wen, Cen, Bai, Qiao, Wang, Wang, Zhu, Liu, Li, Wang, Wen, Huang, Cai, Yu, Li, Hu, Zhang, Zhang, Lin, Yang, Wang, Ai, Zhu, Yi, Chen, Wen, Sun, Zhao, Hu, Zhang, Liu, Zhang, Peng, Tai, Zhang, Liu, Wang, Yan, Ge, Liu, Chu, Zhao, Wang, Zhao, Ren, Wang, Zhang, Gui, Zhao, Li, An, Li, Yuan, Du, Liu, Zhi, Duan, Zhou, Wei, Wang, Luo, Zhang, Sha, Xu, Wu, Ding, Chen, Li, Lin, Ta, Zou, Song, Yang, Tu, Yang, Wu, Zhang, Li, Li, Fan, Qin, Tian, Zhang, Yu, Liang, Kuang, Cheng, Li, Yan, Hu, Ling, Fan, Xia, Zhang, Zhang, Pan, Zou, Zhang, Liu, Wu, Li, Wang, Zhu, Tan, Zhou, Pan, Zhang, Su, Geng, Yan, Tan, Bi, Shen, Yang, Li, Liu, Wang, Li, Wu, Zhang, Duan, Zhang, Liu, Jiang, Yan, Zhang, Wei, Chen, Feng, Yao, Chai, Wang, Zhang, Xu,
  Huang, Wang, Li, Dong, and Tang}]{glm5team2026glm5vibecodingagentic}
GLM-5-Team, Aohan Zeng, Xin Lv, Zhenyu Hou, Zhengxiao Du, Qinkai Zheng, Bin Chen, Da~Yin, Chendi Ge, Chenghua Huang, Chengxing Xie, Chenzheng Zhu, Congfeng Yin, Cunxiang Wang, Gengzheng Pan, Hao Zeng, Haoke Zhang, Haoran Wang, Huilong Chen, and 167 others. 2026.
\newblock \href {https://arxiv.org/abs/2602.15763} {Glm-5: from vibe coding to agentic engineering}.
\newblock \emph{Preprint}, arXiv:2602.15763.

\bibitem[{Gu et~al.(2026)Gu, Dong, Wei, and Huang}]{gu2026minillmonpolicydistillationlarge}
Yuxian Gu, Li~Dong, Furu Wei, and Minlie Huang. 2026.
\newblock \href {https://arxiv.org/abs/2306.08543} {Minillm: On-policy distillation of large language models}.
\newblock \emph{Preprint}, arXiv:2306.08543.

\bibitem[{Hendrycks et~al.(2021)Hendrycks, Burns, Kadavath, Arora, Basart, Tang, Song, and Steinhardt}]{hendrycks2021math}
Dan Hendrycks, Collin Burns, Saurav Kadavath, Akul Arora, Steven Basart, Eric Tang, Dawn Song, and Jacob Steinhardt. 2021.
\newblock \href {https://arxiv.org/abs/2103.03874} {Measuring mathematical problem solving with the {MATH} dataset}.
\newblock \emph{NeurIPS}.

\bibitem[{Hinton et~al.(2015)Hinton, Vinyals, and Dean}]{hinton2015distillingknowledgeneuralnetwork}
Geoffrey Hinton, Oriol Vinyals, and Jeff Dean. 2015.
\newblock \href {https://arxiv.org/abs/1503.02531} {Distilling the knowledge in a neural network}.
\newblock \emph{Preprint}, arXiv:1503.02531.

\bibitem[{Hou et~al.(2026)Hou, Peng, Wang, Ruan, Zhang, Zhou, Gao, Chen, Wang, Yang et~al.}]{hou2026uni}
Wenjin Hou, Shangpin Peng, Weinong Wang, Zheng Ruan, Yue Zhang, Zhenglin Zhou, Mingqi Gao, Yifei Chen, Kaiqi Wang, Hongming Yang, and 1 others. 2026.
\newblock Uni-opd: Unifying on-policy distillation with a dual-perspective recipe.
\newblock \emph{arXiv preprint arXiv:2605.03677}.

\bibitem[{Huang et~al.(2022)Huang, You, Wang, Qian, and Xu}]{kd-from-strong-1}
Tao Huang, Shan You, Fei Wang, Chen Qian, and Chang Xu. 2022.
\newblock \href {https://arxiv.org/abs/2205.10536} {Knowledge distillation from a stronger teacher}.
\newblock \emph{Preprint}, arXiv:2205.10536.

\bibitem[{Jin et~al.(2026)Jin, Min, Yang, Kadhe, Zhou, Wei, Baracaldo, and Lee}]{eopd}
Woogyeol Jin, Taywon Min, Yongjin Yang, Swanand~Ravindra Kadhe, Yi~Zhou, Dennis Wei, Nathalie Baracaldo, and Kimin Lee. 2026.
\newblock \href {https://arxiv.org/abs/2603.07079} {Entropy-aware on-policy distillation of language models}.
\newblock \emph{Preprint}, arXiv:2603.07079.

\bibitem[{Lewkowycz et~al.(2022)Lewkowycz, Andreassen, Dohan, Dyer, Michalewski, Ramasesh, Slone, Anil, Schlag, Gutman-Solo et~al.}]{lewkowycz2022minerva}
Aitor Lewkowycz, Anders Andreassen, David Dohan, Ethan Dyer, Henryk Michalewski, Vinay Ramasesh, Ambrose Slone, Cem Anil, Imanol Schlag, Theo Gutman-Solo, and 1 others. 2022.
\newblock \href {https://arxiv.org/abs/2206.14858} {Solving quantitative reasoning problems with language models}.
\newblock \emph{NeurIPS}.

\bibitem[{Li et~al.(2026{\natexlab{a}})Li, Zuo, He, Zhang, Xiao, Qian, Yu, ang Gao, Yang, Liu, and Ding}]{li2026rethinkingonpolicydistillationlarge}
Yaxuan Li, Yuxin Zuo, Bingxiang He, Jinqian Zhang, Chaojun Xiao, Cheng Qian, Tianyu Yu, Huan ang Gao, Wenkai Yang, Zhiyuan Liu, and Ning Ding. 2026{\natexlab{a}}.
\newblock \href {https://arxiv.org/abs/2604.13016} {Rethinking on-policy distillation of large language models: Phenomenology, mechanism, and recipe}.
\newblock \emph{Preprint}, arXiv:2604.13016.

\bibitem[{Li et~al.(2025)Li, Yue, Xu, Jiang, Niu, Lin, Ramasubramanian, and Poovendran}]{li2025learnability}
Yuetai Li, Xiang Yue, Zhangchen Xu, Fengqing Jiang, Luyao Niu, Bill~Yuchen Lin, Bhaskar Ramasubramanian, and Radha Poovendran. 2025.
\newblock \href {https://doi.org/10.18653/v1/2025.findings-acl.1301} {Small models struggle to learn from strong reasoners}.
\newblock In \emph{Findings of the Association for Computational Linguistics: ACL 2025}, pages 25366--25394, Vienna, Austria. Association for Computational Linguistics.

\bibitem[{Li et~al.(2026{\natexlab{b}})Li, Xi, Chen, Wang, Jiang, Shen, Song, Wei, and Lian}]{li2026rolereasoningpatternsgeneralization}
Zhaoyi Li, Xiangyu Xi, Zhengyu Chen, Wei Wang, Gangwei Jiang, Ranran Shen, Linqi Song, Ying Wei, and Defu Lian. 2026{\natexlab{b}}.
\newblock \href {https://arxiv.org/abs/2604.01702} {On the role of reasoning patterns in the generalization discrepancy of long chain-of-thought supervised fine-tuning}.
\newblock \emph{Preprint}, arXiv:2604.01702.

\bibitem[{Lu and Lab(2025)}]{lu2025onpolicydistillation}
Kevin Lu and Thinking~Machines Lab. 2025.
\newblock \href {https://doi.org/10.64434/tml.20251026} {On-policy distillation}.
\newblock \emph{Thinking Machines Lab: Connectionism}.
\newblock Https://thinkingmachines.ai/blog/on-policy-distillation.

\bibitem[{Luo et~al.(2026)Luo, Chuang, Wang, Xu, Han, Zhang, and Braverman}]{opd-var-1}
Feng Luo, Yu-Neng Chuang, Guanchu Wang, Zicheng Xu, Xiaotian Han, Tianyi Zhang, and Vladimir Braverman. 2026.
\newblock \href {https://arxiv.org/abs/2604.08527} {Demystifying opd: Length inflation and stabilization strategies for large language models}.
\newblock \emph{Preprint}, arXiv:2604.08527.

\bibitem[{{MAA}(2026{\natexlab{a}})}]{MAA_AIME}
{MAA}. 2026{\natexlab{a}}.
\newblock American invitational mathematics examination ({AIME}).
\newblock \url{https://www.maa.org/math-competitions}.

\bibitem[{{MAA}(2026{\natexlab{b}})}]{MAA_AMC}
{MAA}. 2026{\natexlab{b}}.
\newblock American mathematics competitions ({AMC}).
\newblock \url{https://www.maa.org/math-competitions}.

\bibitem[{Mirzadeh et~al.(2019)Mirzadeh, Farajtabar, Li, Levine, Matsukawa, and Ghasemzadeh}]{mirzadeh2020improved}
Seyed-Iman Mirzadeh, Mehrdad Farajtabar, Ang Li, Nir Levine, Akihiro Matsukawa, and Hassan Ghasemzadeh. 2019.
\newblock \href {https://arxiv.org/abs/1902.03393} {Improved knowledge distillation via teacher assistant}.
\newblock \emph{Preprint}, arXiv:1902.03393.

\bibitem[{Schulman et~al.(2017)Schulman, Wolski, Dhariwal, Radford, and Klimov}]{schulman2017ppo}
John Schulman, Filip Wolski, Prafulla Dhariwal, Alec Radford, and Oleg Klimov. 2017.
\newblock \href {https://arxiv.org/abs/1707.06347} {Proximal policy optimization algorithms}.
\newblock \emph{Preprint}, arXiv:1707.06347.

\bibitem[{Shao et~al.(2024)Shao, Wang, Zhu, Xu, Song, Bi, Zhang, Zhang, Li, Wu, and Guo}]{grpo}
Zhihong Shao, Peiyi Wang, Qihao Zhu, Runxin Xu, Junxiao Song, Xiao Bi, Haowei Zhang, Mingchuan Zhang, Y.~K. Li, Y.~Wu, and Daya Guo. 2024.
\newblock \href {https://arxiv.org/abs/2402.03300} {Deepseekmath: Pushing the limits of mathematical reasoning in open language models}.
\newblock \emph{Preprint}, arXiv:2402.03300.

\bibitem[{Wang et~al.(2026)Wang, Lee, Lee, Lin, Dai, Chen, Chen, Yang, Liu, Shoeybi, Catanzaro, and Ping}]{nemotron}
Boxin Wang, Chankyu Lee, Nayeon Lee, Sheng-Chieh Lin, Wenliang Dai, Yang Chen, Yangyi Chen, Zhuolin Yang, Zihan Liu, Mohammad Shoeybi, Bryan Catanzaro, and Wei Ping. 2026.
\newblock \href {https://arxiv.org/abs/2512.13607} {Nemotron-cascade: Scaling cascaded reinforcement learning for general-purpose reasoning models}.
\newblock \emph{Preprint}, arXiv:2512.13607.

\bibitem[{Yang et~al.(2025)Yang, Li, Yang, Zhang, Hui, Zheng, Yu, Gao, Huang, Lv, Zheng, Liu, Zhou, Huang, Hu, Ge, Wei, Lin, Tang, Yang, Tu, Zhang, Yang, Yang, Zhou, Zhou, Lin, Dang, Bao, Yang, Yu, Deng, Li, Xue, Li, Zhang, Wang, Zhu, Men, Gao, Liu, Luo, Li, Tang, Yin, Ren, Wang, Zhang, Ren, Fan, Su, Zhang, Zhang, Wan, Liu, Wang, Cui, Zhang, Zhou, and Qiu}]{qwen3}
An~Yang, Anfeng Li, Baosong Yang, Beichen Zhang, Binyuan Hui, Bo~Zheng, Bowen Yu, Chang Gao, Chengen Huang, Chenxu Lv, Chujie Zheng, Dayiheng Liu, Fan Zhou, Fei Huang, Feng Hu, Hao Ge, Haoran Wei, Huan Lin, Jialong Tang, and 41 others. 2025.
\newblock \href {https://arxiv.org/abs/2505.09388} {Qwen3 technical report}.
\newblock \emph{Preprint}, arXiv:2505.09388.

\bibitem[{Yang et~al.(2026{\natexlab{a}})Yang, Liu, Xie, Yang, Yang, and Lin}]{gopd}
Wenkai Yang, Weijie Liu, Ruobing Xie, Kai Yang, Saiyong Yang, and Yankai Lin. 2026{\natexlab{a}}.
\newblock \href {https://arxiv.org/abs/2602.12125} {Learning beyond teacher: Generalized on-policy distillation with reward extrapolation}.
\newblock \emph{Preprint}, arXiv:2602.12125.

\bibitem[{Yang et~al.(2026{\natexlab{b}})Yang, Liu, Chen, Dai, Wang, Lin, Lee, Chen, Jiang, He, Pi, Lam, Lee, Bukharin, Shoeybi, Catanzaro, and Ping}]{Nemotron-Cascade-2}
Zhuolin Yang, Zihan Liu, Yang Chen, Wenliang Dai, Boxin Wang, Sheng-Chieh Lin, Chankyu Lee, Yangyi Chen, Dongfu Jiang, Jiafan He, Renjie Pi, Grace Lam, Nayeon Lee, Alexander Bukharin, Mohammad Shoeybi, Bryan Catanzaro, and Wei Ping. 2026{\natexlab{b}}.
\newblock Nemotron-cascade 2: Post-training llms with cascade rl and multi-domain on-policy distillation.

\bibitem[{Yu et~al.(2025)Yu, Zhang, Zhu, Yuan, Zuo, Yue, Dai, Fan, Liu, Liu, Liu, Lin, Lin, Ma, Sheng, Tong, Zhang, Zhang, Zhang, Zhu, Zhu, Chen, Chen, Wang, Yu, Song, Wei, Zhou, Liu, Ma, Zhang, Yan, Qiao, Wu, and Wang}]{dapo}
Qiying Yu, Zheng Zhang, Ruofei Zhu, Yufeng Yuan, Xiaochen Zuo, Yu~Yue, Weinan Dai, Tiantian Fan, Gaohong Liu, Lingjun Liu, Xin Liu, Haibin Lin, Zhiqi Lin, Bole Ma, Guangming Sheng, Yuxuan Tong, Chi Zhang, Mofan Zhang, Wang Zhang, and 16 others. 2025.
\newblock \href {https://arxiv.org/abs/2503.14476} {Dapo: An open-source llm reinforcement learning system at scale}.
\newblock \emph{Preprint}, arXiv:2503.14476.

\bibitem[{Yu et~al.(2026)Yu, Li, Si, Zhang, Xu, Wang, Dong, Tuo, Zeng, Feng, Wang, Shi, Hu, Yue, Wang, and Yan}]{yu2026dopddualonpolicydistillation}
Xinlei Yu, Gen Li, Qingyi Si, Guibin Zhang, Yuqi Xu, Congcong Wang, Shuai Dong, Kaiwen Tuo, Xiangyu Zeng, Kaituo Feng, Qunzhong Wang, Yang Shi, Xiaobin Hu, Xiangyu Yue, Jiaqi Wang, and Shuicheng Yan. 2026.
\newblock \href {https://arxiv.org/abs/2606.30626} {Dopd: Dual on-policy distillation}.
\newblock \emph{Preprint}, arXiv:2606.30626.

\bibitem[{Yue et~al.(2025)Yue, Chen, Lu, Zhao, Wang, Yue, Song, and Huang}]{RL-no-up}
Yang Yue, Zhiqi Chen, Rui Lu, Andrew Zhao, Zhaokai Wang, Yang Yue, Shiji Song, and Gao Huang. 2025.
\newblock \href {https://arxiv.org/abs/2504.13837} {Does reinforcement learning really incentivize reasoning capacity in llms beyond the base model?}
\newblock \emph{Preprint}, arXiv:2504.13837.

\bibitem[{Zhu et~al.(2026)Zhu, Ye, Lu, Shi, and Liu}]{zhu2026facesonpolicydistillationpitfalls}
Siqi Zhu, Xuyan Ye, Hongyu Lu, Weiye Shi, and Ge~Liu. 2026.
\newblock \href {https://arxiv.org/abs/2605.11182} {The many faces of on-policy distillation: Pitfalls, mechanisms, and fixes}.
\newblock \emph{Preprint}, arXiv:2605.11182.

\end{thebibliography}

\appendix

\section{Experiment Configurations}
We implement the training pipeline using {verl}.
For the OPD pipeline, policy optimization uses a total batch size of $128$, 1 rollout for each prompt. 
For the GRPO baselines, the group advantage is calculated across a size of $128 \times 8$ rollouts. Optimization is limited to a single epoch per rollout batch to eliminate off-policy bias.
The maximum sequence length is set to $16,384$ tokens during training. 
Rollouts are generated using stochastic sampling with a temperature of $\tau = 1.0$ and $\text{top-}p = 1.0$.

For the avg@32 metric, we utilized $N=40$ candidate responses per problem for better estimation. To obtain a stable estimation, we perform $M=10$ independent trials. In each trial, we randomly sample a subset of $k=32$ responses from the candidate pool and calculate the correctness of the $j$-th response as $\mathbb{I}_j \in \{0, 1\}$. The final score is the average across all trials, formulated as:
\begin{equation}
\text{avg@}32 = \frac{1}{M} \sum_{i=1}^{M} \left( \frac{1}{k} \sum_{j \in \mathcal{S}_i} \mathbb{I}_j \right)
\end{equation}


\end{document}